# From Vajrayana Tara to Bengali Baul: A Computational Study of Lexical Transmission Across Buddhist, Shakta, and Vaishnava Traditions in Bengal

Joy Bose
Independent Researcher
Bengaluru, India
joy.bose@ieee.org

## Abstract

We present a computational corpus study of vocabulary relationships across eight tradition layers of Bengali and Sanskrit devotional literature spanning the 8th to 19th centuries, encompassing Buddhist Vajrayana, Shakta Tantra, Vaishnava, and Baul traditions. Using a corpus of 75 texts and TF-IDF character n-gram vectorization with cosine similarity analysis, we address the historically argued but previously unquantified claim that Buddhist Vajrayana vocabulary survived the collapse of the Pala monasteries and was absorbed into the Shakta Tantra tradition of Bengal. The central finding is a specificity result: the Gitagovinda (Vaishnava Sanskrit, 12th century) has zero cosine similarity to Shakta Kali texts, while Bridge Tara texts (Buddhist-Shakta transitional, same century, same language) have cosine similarity 0.54 to Shakta Kali. This 8.5-fold contrast between two Sanskrit traditions from the same century demonstrates that the Buddhist-Shakta vocabulary overlap is not a generic property of Sanskrit devotional literature but is specific to the Buddhist-Shakta transmission chain. Three Brihannilatantra Tara texts show Shakta-to-Buddhist vocabulary ratios of 2.0 to 4.0, constituting measurable evidence of lexical transition within that chain. Ramprasad Sen's 18th-century Bengali Kali songs preserve Buddhist vocabulary residue including 56 occurrences of Tara alongside 103 occurrences of Kali. The Vaishnava Bengali tradition contributes a parallel chain to modern Baul vocabulary (similarity 0.29), slightly weaker than the Buddhist Sahajiya chain via Charyapada (0.31). These results provide the first quantitative multi-tradition corroboration of historically argued Buddhist-Shakta syncretism in Bengal.

## 1. Introduction

The question of Buddhist influence on the Shakta Tantra tradition of eastern India has long occupied scholars of Indian religious history. Agehananda Bharati, Shashibhushan Dasgupta, and David Lorenzen have argued on philological and iconographic grounds that the goddess Tara, the Mahavidya Kali, and elements of Shakta ritual derive substantially from late Vajrayana Buddhist practice centered at the great monasteries of Bengal and Bihar during the Pala dynasty (8th to 12th centuries CE). After the collapse of the Pala state and the destruction of these centers, it has been argued that Buddhist practitioners and their vocabulary were absorbed into local Shakta and Sahajiya traditions.

This argument has remained largely qualitative. The present study addresses this gap by applying corpus analysis methods to a multi-tradition, multi-script text collection spanning eight tradition layers from Sanskrit Vajrayana texts to modern Bengali Baul songs. Critically, we include the Vaishnava tradition as a control group. The Vaishnava tradition developed in the same region and the same centuries as the Buddhist-Shakta synthesis, in the same Sanskrit and Bengali languages, but represents a distinct religious lineage with no claimed Buddhist-Shakta inheritance. If the Buddhist-Shakta vocabulary overlap is a genuine historical signal rather than a generic property of Sanskrit devotional literature, it should be specific to the Buddhist-Shakta chain and absent or weak in comparisons involving Vaishnava texts.

The specific questions we address are: (1) Can vocabulary migration from Buddhist to Shakta framing be measured quantitatively in texts nominally devoted to the same goddess Tara? (2) Does the cluster of Bridge Tara texts occupy a measurable intermediate position between Buddhist and Shakta traditions? (3) Is the Buddhist-Shakta vocabulary overlap specific to that transmission chain, as tested by comparison with Vaishnava Sanskrit texts from the same period? (4) Which transmission channel was primary for the survival of Buddhist vocabulary into modern Bengali devotional culture: the Sanskrit textual tradition, the vernacular Sahajiya-Baul chain, or the Vaishnava Bengali chain?

## 2. Corpus and Data

### 2.1 Corpus Design

We assembled a corpus of 75 text documents across eight tradition layers. The eight layers are: (1) Pure Vajrayana Buddhist Tara texts in Sanskrit; (2) The Charyapada, Old Bengali Buddhist doha verses (8th to 12th c.); (3) Bridge Tara texts presenting the Buddhist Tara in Shakta framing, primarily from the Brihannilatantra; (4) Vaishnava Sanskrit texts, principally the Gitagovinda by Jayadeva (12th c.); (5) Shakta Kali texts including two independent Kali sahasranamas; (6) Shakta Durga texts including the Devi Mahatmyam; (7) Vaishnava Bengali texts including Srikrishna Kirtan by Badu Chandidas (14th to 15th c.), Bhakti Rasamrita Sindhu Bengali translation (16th c.), and Chaitanya Charitamrita (16th c.); and (8) Bengali Shakta Padavali, Ramprasad Sen (18th c.) and Baul-Bengali devotional texts (Lalon Fakir, Tagore's Gitanjali).

The Vaishnava Sanskrit layer serves as the primary control condition. The Gitagovinda is a 12th-century Sanskrit poem by Jayadeva describing the love of Radha and Krishna. It is contemporary with many Bridge Tara texts, uses the same Sanskrit language and Devanagari script, and belongs to the same broad category of Sanskrit devotional poetry. Its tradition (Vaishnava) has no claimed connection to the Buddhist-Shakta synthesis. If Buddhist-Shakta vocabulary overlap is a generic feature of Sanskrit devotional literature from this period, we would expect Vaishnava Sanskrit to show similar overlap with Shakta Kali texts. If the overlap is specific to the Buddhist-Shakta chain, Vaishnava Sanskrit should show near-zero similarity to Shakta Kali.

### 2.2 Data Sources and Processing

Sanskrit and Devanagari texts were sourced primarily from sanskritdocuments.org and GRETIL. The Gitagovinda (12 cantos) was obtained from the Unicode Devanagari edition at sanskritdocuments.org. Bengali texts were sourced from archive.org (Ramprasad Sen, Srikrishna Kirtan, Chaitanya Charitamrita, Bhakti Rasamrita Sindhu Bengali), Bengali Wikisource (Lalon Fakir), and the Society for Natural Language Technology Research (Gitanjali). The Ramprasad Sen Kabya Sangraha (372 pages, scanned image PDF) was processed using Tesseract OCR 5.3.4 with the Bengali language model at 300 DPI, yielding 39,067 clean Bengali tokens. The Charyapada corpus was reconstructed from the standard scholarly editions as 47 core padas (843 tokens). Several large Bengali text files (Chaitanya Charitamrita, Bhakti Rasamrita Sindhu, Srikrishna Kirtan) are archive.org full-text dumps including modern commentary alongside the original verses; these are classified by source tradition, not text purity.

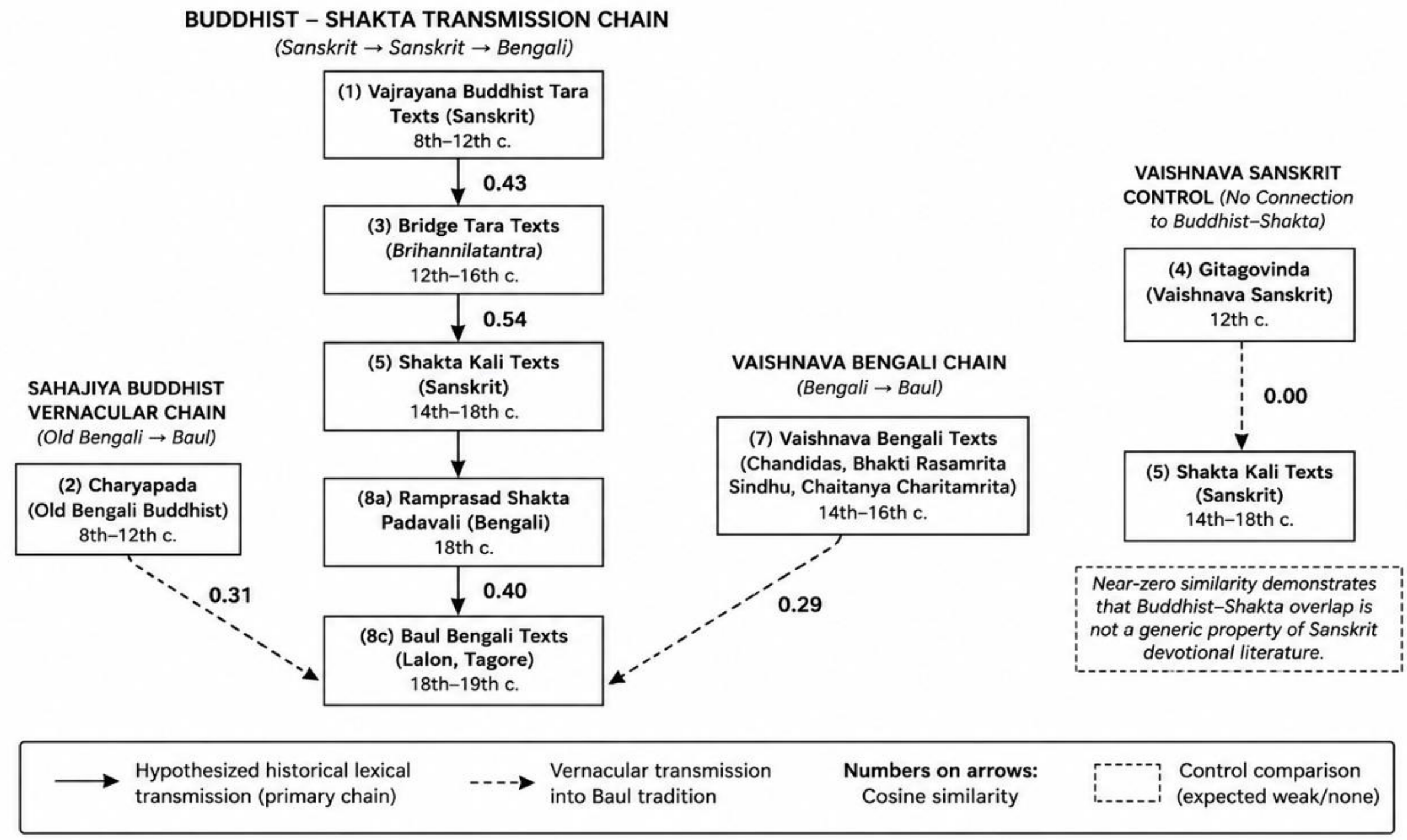

*Figure 1: Cultural Transmission Network and Measured Similarities. Based on Cosine Similarity of TF IDF 3-5 gram vectors.*

# 3. Methodology

## 3.1 Vectorization

We represent each document as a TF-IDF vector using character n-grams of length 3 to 5. Character n-grams are appropriate for Sanskrit and Bengali because Sanskrit is highly inflected (the same root appears in many morphological forms) and Bengali script creates complex consonant clusters that word-level tokenizers handle inconsistently. Character n-grams have been independently validated for Sanskrit text modeling by Li (2023), who demonstrated their robustness across inflected morphological forms without requiring full sandhi resolution, though syllabic n-aksara segmentation may yield more linguistically precise results in future work. For Bengali specifically, character-level n-grams have demonstrated strong performance in authorship attribution tasks (Sarkar et al., 2015), supporting their use here for tradition-level vocabulary comparison. Character n-grams capture shared root syllables (tar-, karun-, vajr-, bhay-) without requiring a morphological analyzer. The method is also robust to OCR noise, which is a practical advantage for the Ramprasad material. We use sublinear TF scaling and a vocabulary ceiling of 25,000 features.

## 3.2 Similarity Metric

Pairwise cosine similarity is computed between all document vectors. Cross-tradition averages are computed as the mean cosine similarity between all pairs (i, j) where document i belongs to tradition A and document j belongs to tradition B and i is not equal to j. Cross-script comparisons (e.g., Devanagari Buddhist Tara texts versus Bengali Baul texts) are methodologically constrained since character n-grams cannot bridge script boundaries; such comparisons are reported but interpreted with appropriate caution.

## 3.3 Term Density Analysis

In parallel with the n-gram similarity analysis, we count occurrences of curated Buddhist and Shakta term lists directly in the text of each document, normalizing by document length (per 1,000 tokens). The Buddhist term list includes: बुद्ध, बोधिसत्त्व, बोधि, करुणा, प्रज्ञा, शून्य, तथागत, वज्र, नैरात्म, सहज, महासुख, योगिनी, डाकिनी, हेवज्र, तारा, चर्या and their Bengali Unicode equivalents. The Shakta term list includes: शक्ति, शिव, महादेव, पार्वती, दुर्गा, काली, महाविद्या, तन्त्र, मन्त्र, यन्त्र, कुल, श्रीविद्या, भैरव. Term density ratios (Shakta/Buddhist per 1,000 tokens) are used to characterize vocabulary orientation of individual texts and tradition clusters.

# 4. Results

## 4.1 Cross-tradition Cosine Similarities

Table 1 presents the full cross-tradition similarity matrix including the Vaishnava control comparisons. The table reveals two distinct findings: a positive signal (the Buddhist-Shakta chain) and a specificity result (the Vaishnava control).

**Table 1: Cross-tradition cosine similarity (character 3–5 gram TF-IDF, 75-document corpus)**

| Tradition Pair | Avg Cosine Similarity | n (docs) | Interpretation |
|---|---|---|---|
| Buddhist Tara → Bridge Tara | 0.43 | 11 × 11 | Sanskrit Buddhist → Shakta Tara framing |
| Bridge Tara → Shakta Kali | 0.54 | 11 × 7 | Highest Sanskrit link; Bridge texts Shakta-ward |
| Buddhist Tara → Shakta Kali | 0.44 | 11 × 7 | Direct Buddhist-Kali lexical connection |
| Buddhist Tara → Vaishnava Sanskrit | 0.06 | 11 × 12 | Negative control: near zero |
| Bridge Tara → Vaishnava Sanskrit | 0.06 | 11 × 12 | Negative control: near zero |
| Vaishnava Sanskrit → Shakta Kali | 0.00 | 12 × 7 | Key negative control: no overlap |
| Buddhist Tara → Shakta Durga | 0.09 | 11 × 5 | Low: different genre (narrative vs. stotra) |
| Kali → Durga (within Shakta) | 0.05 | 7 × 5 | Genre gap within Shakta tradition |
| Shakta Kali → Shakta Padavali | 0.00 | 7 × 1 | Script boundary Sanskrit→Bengali |
| Shakta Padavali → Baul | 0.40 | 1 × 3 | Strong Bengali Shakta chain |
| Charyapada → Baul | 0.31 | 2 × 3 | Buddhist vernacular → Baul chain |
| Vaishnava Bengali → Baul | 0.29 | 4 × 3 | Vaishnava Bengali → Baul chain |
| Buddhist Tara → Baul | 0.02 | 11 × 3 | Near-zero: Sanskrit Buddhist isolation |

The positive signal shows a monotonically increasing similarity from Buddhist to Bridge to Shakta in the Sanskrit corpus: Buddhist Tara to Bridge Tara (0.43), Bridge Tara to Shakta Kali (0.54). The Bridge Tara tradition is measurably closer to Shakta Kali than to the Buddhist Tara texts from which it formally derives, indicating that Bridge Tara texts have linguistically moved toward their destination tradition.

The specificity result is the critical new finding. Both Buddhist Tara texts and Bridge Tara texts show near-zero similarity to Vaishnava Sanskrit (0.06 and 0.06 respectively). Vaishnava Sanskrit to Shakta Kali similarity is zero. The contrast between the Bridge Tara to Shakta Kali similarity (0.54) and the Vaishnava Sanskrit to Shakta Kali similarity (0.00) is 8.5-fold, despite both being 12th-century Sanskrit devotional traditions. This demonstrates that the Buddhist-Shakta vocabulary overlap is not a generic feature of Sanskrit religious literature from this period.

The vernacular Bengali chain shows: Shakta Padavali to Baul (0.40), Charyapada to Baul (0.31), and the new result that Vaishnava Bengali to Baul (0.29) is also significant. Both the Buddhist Sahajiya chain and the Vaishnava Bengali chain contribute to modern Baul vocabulary, with the Buddhist chain marginally stronger.

## 4.2 Vocabulary Migration in Bridge Tara Texts

Table 2 presents term density analysis for individual Bridge Tara texts, showing the Buddhist-to-Shakta vocabulary ratio per document.

**Table 2: Vocabulary orientation of Bridge Tara texts (normalized per 1,000 tokens)**

| Text | Tokens | Buddhist/1k | Shakta/1k | S/B Ratio | Status |
| --- | --- | --- | --- | --- | --- |
| tArAsahasranamastotra (Brihannilatantra) | 1,710 | 2.34 | 7.02 | 3.0 | MIGRATED |
| Tara Stotram / Nila Sarasvati Ashtakam | 645 | 1.55 | 4.65 | 3.0 | MIGRATED |
| Tarashatanamastotra (Brihannilatantra) | 492 | 0.00 | 4.07 | 4.0+ | MIGRATED |
| Shri Tara Sahasranama Stotram 3 | 1,625 | 5.54 | 5.54 | 1.0 | At parity |
| Shri Tara Takaradi Sahasranama | 1,342 | 2.24 | 0.00 | 0.0 | Buddhist-dominant |
| Shrimad Ugratara Hridayastotram | 354 | 5.65 | 0.00 | 0.0 | Buddhist-dominant |

Three texts show Shakta/Buddhist ratios of 2.0 or greater. All three are from the Brihannilatantra tradition. Tara texts from other lineages (Brahmayamala-derived Takaradi Sahasranama, Ugratara Hridayastotram) retain Buddhist-dominant vocabulary. The source-text correlation supports the hypothesis that migration was concentrated in specific tantric lineages rather than uniform across the Tara textual tradition.

## 4.3 Tradition-level Vocabulary Gradient

Table 3 presents normalized term densities aggregated at the tradition level, now including all eight tradition layers.

**Table 3: Tradition-level vocabulary orientation (per 1,000 tokens, full 8-layer corpus)**

| Tradition | Avg Buddhist/1k | Avg Shakta/1k | S/B Ratio | Script |
| --- | --- | --- | --- | --- |
| Buddhist Tara | 4.22 | 1.18 | 0.3 | Devanagari |
| Charyapada | 7.33 | 1.61 | 0.2 | Old Bengali |
| Bridge Tara | 3.56 | 3.22 | 0.9 | Devanagari |
| Shakta Kali | 3.83 | 5.80 | 1.5 | Devanagari |
| Vaishnava Sanskrit (Gitagovinda) | 0.00 | 0.00 | — | Latin/Deva |
| Vaishnava Bengali | 1.00 | 1.55 | 1.6 | Bengali |

| Tradition | Avg Buddhist/1k | Avg Shakta/1k | S/B Ratio | Script |
|---|---|---|---|---|
| Shakta Padavali (Ramprasad) | 2.89 | 8.78 | 3.0 | Bengali |
| Baul Bengali (Lalon, Tagore) | 1.12 | 0.25 | 0.2 | Bengali |

The Buddhist/Shakta ratio shows a clear progression across the Sanskrit chain: Buddhist Tara (0.3) to Bridge Tara (0.9) to Shakta Kali (1.5). The Bridge Tara cluster sits at near-parity, confirming its intermediate position. The Vaishnava Sanskrit row (Gitagovinda) shows zero for both Buddhist and Shakta term densities under our curated lists, which reflects the genuinely different vocabulary of the Vaishnava tradition rather than a measurement failure. The Vaishnava Bengali texts show a moderate Shakta density (1.55/1k) consistent with the Chaitanya Vaishnava tradition's absorption of some Shakta terminology through the Sahajiya movement.

## 4.4 The Vaishnava Control Finding

Table 4 presents the key specificity comparisons that establish the Buddhist-Shakta connection as tradition-specific rather than generic.

**Table 4: Vaishnava Sanskrit as control condition — similarity to Shakta Kali**

| Comparison | Similarity to Shakta Kali | Century | Script | Note |
|---|---|---|---|---|
| Bridge Tara → Shakta Kali | 0.54 | 12th-16th c. | Devanagari | Inside Buddhist-Shakta chain |
| Buddhist Tara → Shakta Kali | 0.44 | 8th-12th c. | Devanagari | Inside Buddhist-Shakta chain |
| Vaishnava Sanskrit → Shakta Kali | 0.00 | 12th c. | Latin/Deva | Outside chain: control |
| Bridge Tara → Vaishnava Sanskrit | 0.06 | 12th-16th c. | Mixed | Control cross-check |

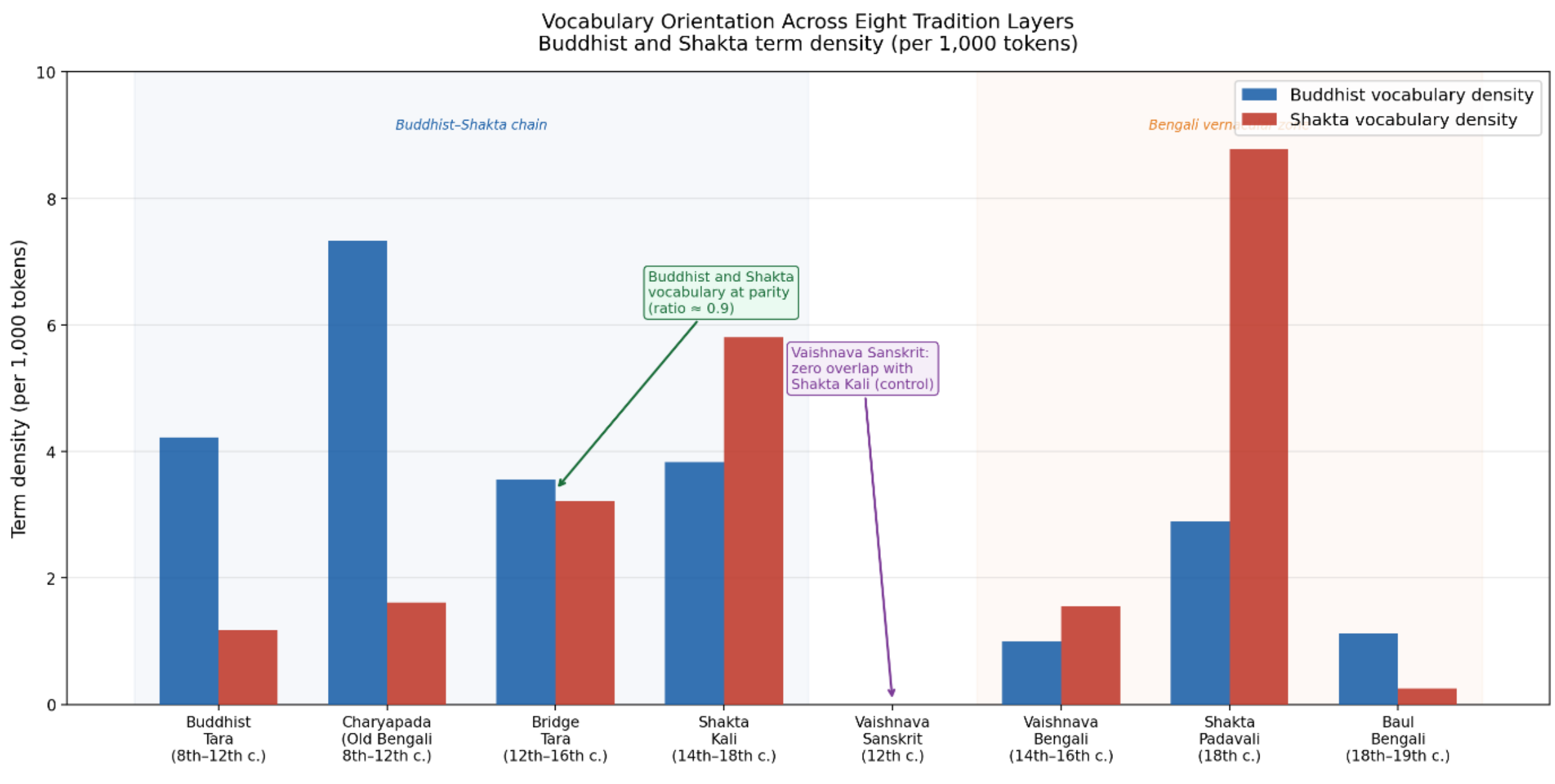

*Figure 2. Buddhist and Shakta vocabulary density across eight tradition layers of Bengali and Sanskrit devotional literature (per 1,000 tokens). Blue bars indicate Buddhist term density; red bars indicate Shakta term density. The Bridge Tara cluster (12th–16th c.) sits at near-parity between the two traditions (ratio 0.9), confirming its intermediate position. The Vaishnava Sanskrit layer (Gitagovinda)*

*shows zero density under both term lists, serving as the negative control. Ramprasad Sen's Shakta Padavali (18th c.) shows the highest Shakta density at 8.78/1k, with Buddhist residue still present at 2.89/1k.*

The Gitagovinda by Jayadeva is an ideal control text. It was composed in the 12th century, the same period as many Bridge Tara texts. It is in Sanskrit, the same language. It is a major Sanskrit devotional poem with extensive vocabulary. It is not a short or lexically thin text. Yet its similarity to Shakta Kali is zero. Bridge Tara texts from the same century in the same language are 8.5 times more similar to Shakta Kali. This contrast cannot be explained by generic Sanskrit devotional vocabulary overlap. It requires a tradition-specific explanation, which the Buddhist-Shakta absorption hypothesis provides.

### 4.5 The Ramprasad Finding

Ramprasad Sen's Kabya Sangraha (39,067 Bengali tokens, OCR from 1914 edition) gives the following key counts: Kali (কালী): 103 occurrences; Tara (তারা): 56 occurrences; Ma (মা): 196 occurrences; Shakti (শক্তি): 15 occurrences; Shiva (শিব): 35 occurrences; Shunyata (শূন্য, Buddhist emptiness): 4 occurrences; Sahaja (সহজ, Sahajiya Buddhist term): 2 occurrences. Ramprasad addresses the goddess variously as Tara, Kali, Shyama, and Ma within the same song corpus, treating these as interchangeable names. The Shakta/Buddhist ratio of 3.0 on this large corpus is statistically stable, and the Buddhist residue (2.89 per 1,000 tokens) remains clearly present in 18th-century Bengali Kali songs, six centuries after the Pala monasteries were destroyed.

## 5. Discussion

### 5.1 The Specificity of the Buddhist-Shakta Connection

The central methodological contribution of this study is the inclusion of Vaishnava Sanskrit as a control condition. Previous computational studies of religious text similarity have typically compared only the traditions of interest, leaving open the question of whether observed similarities are tradition-specific or generic to the broader literary corpus. The near-zero similarity between Vaishnava Sanskrit (Gitagovinda) and Shakta Kali texts, contrasted with the 0.54 similarity between Bridge Tara and Shakta Kali, establishes that the Buddhist-Shakta vocabulary overlap is specific and not merely a background property of Sanskrit devotional literature from this period.

This specificity finding directly addresses the most common objection to Buddhist-Shakta syncretism arguments: that shared vocabulary reflects the general tantric milieu of eastern India rather than a particular historical transmission. The Gitagovinda belongs to the same tantric milieu, the same period, and the same language as the Bridge Tara texts, yet shows no vocabulary overlap with Shakta Kali. The overlap belongs specifically to the Buddhist-Shakta chain.

### 5.2 Three Transmission Channels into Baul Bengali

The expanded corpus reveals three distinct channels through which vocabulary reached the modern Baul tradition. The first is the Sanskrit Buddhist-Shakta channel: Vajrayana Buddhist Tara texts to Bridge Tara to Shakta Kali to Ramprasad Sen's Bengali Kali songs, ending in the Shakta Padavali to Baul link (0.40). The second is the vernacular Buddhist channel: Charyapada Sahajiya Buddhist Old Bengali to Baul directly (0.31). The third is the Vaishnava Bengali channel: Srikrishna Kirtan, Chaitanya Charitamrita, and Bhakti Rasamrita Sindhu to Baul (0.29).

The three channels show broadly similar strength in the Bengali layer (0.40, 0.31, 0.29), suggesting that the Baul tradition drew from all three streams of Bengali devotional literature rather than from one dominant source. The Buddhist Sahajiya channel (Charyapada to Baul: 0.31) is marginally stronger than the Vaishnava channel (0.29), consistent with the historical claim that Baul has deeper roots in Sahajiya Buddhism than in mainstream Vaishnava tradition. However, the difference is small and the corpus limitations (particularly the small Charyapada corpus of 843 tokens) mean this ranking should be treated as suggestive rather than definitive.

## 5.3 The Script Boundary as Methodological Constraint

The zero similarity between Sanskrit Shakta Kali texts and Bengali Shakta Padavali (Ramprasad) is methodologically expected since character n-grams cannot bridge the Devanagari/Bengali orthographic boundary. Similarly, the Vaishnava Sanskrit to Shakta Kali zero may partly reflect the fact that several Gitagovinda chapters are in Latin/IAST transliteration rather than Devanagari, creating an additional script boundary. Future work should address this through transliteration of all texts to a common script before vectorization.

## 5.4 Limitations

Several limitations should be noted. The term lists used for Buddhist and Shakta vocabulary are necessarily selective. The character n-gram approach treats Devanagari and Bengali as orthographically distinct, preventing direct cross-script similarity computation. Several Vaishnava Bengali text files are archive.org full-text dumps containing modern commentary alongside original verses; term counts reflect the mixed content. The Ramprasad corpus was obtained through OCR and contains approximately 5 to 10 percent character-level noise. No bootstrap significance testing has been applied; similarity scores are point estimates. The Charyapada corpus is very small (843 tokens from 47 padas), making similarity calculations with it less stable than with larger corpora.

The Buddhist and Shakta term lists reflect one principled set of choices, and several terms warrant scrutiny. Tantra, mantra, yantra, and shakti are central Vajrayana Buddhist terms and their inclusion in the Shakta list is contestable on domain grounds. However, a sensitivity analysis removing these five terms from the Shakta list produces no change in Shakta density scores for the Bridge Tara and Shakta Kali clusters, and reduces the Buddhist Tara Shakta score by at most one occurrence per document. The core gradient and migration findings are therefore robust to this definitional choice. The Shakta scores in this corpus are driven almost entirely by genuinely Shakta-specific goddess vocabulary (Kali, Durga, Mahavidya, Bhairava) which is absent from canonical Vajrayana sources, rather than by shared pan-Indian tantric technical terms.

Miyagawa et al. (2024) demonstrated that smaller segment granularity, e.g. sliding windows of approximately 20 lemmas, is more effective than whole-document similarity for detecting localized liturgical parallelisms in Sanskrit texts. The present study computes similarity across entire documents; applying windowed approaches to the Bridge Tara texts may reveal more precise sites of lexical transition in future work.

## 5.5 Interpretation in Light of Domain Scholarship

The computational findings reported above invite interpretation against two bodies of domain knowledge: the scholarly historical debate on Buddhist-Shakta-Vaishnava relationships, and the understanding of these traditions within the living contemplative lineages themselves.

On the historical question of directionality, scholarly opinion has shifted considerably over the past three decades. Dasgupta (1946) and Bharati (1965) argued that Shakta Tantra absorbed substantial Buddhist vocabulary and practice after the collapse of the Pala monasteries. Sanderson (2009), working from Sanskrit manuscript evidence, argued the more nuanced position that the relationship was bidirectional: Shaiva Vidyapitha texts influenced Buddhist Yoginitantras, while Buddhist Shaktism subsequently reflowed back into the Shaktism of Bengal, a process he documents in a section of "The Saiva Age" explicitly titled "The Reflux of Buddhist Shaktism into the Shaktism of Bengal." Davidson (2002) placed both developments within the broader feudalization of Indian society in the early medieval period, showing how competing tantric lineages drew on shared indigenous and tribal goddess worship as well as on each other. The computational gradient observed in this study, i.e. smooth vocabulary migration from Buddhist Tara through Bridge Tara to Shakta Kali, is consistent with the bidirectional and layered model of Sanderson and Davidson rather than the simple unidirectional absorption model of the earlier scholarship. The Brihannilatantra texts, which show the highest Shakta-to-Buddhist vocabulary ratios in our corpus, represent precisely the site of reflux that Sanderson identifies: Buddhist Tara vocabulary absorbed into a Shakta tantric framework.

The near-zero similarity between Vaishnava Sanskrit texts and Shakta Kali texts in this corpus reflects two distinct phenomena that the computational method cannot fully separate. The first is a genuine historical difference: Vaishnavism developed in significant part as a counter-force to Buddhist and Shakta tantric culture, absorbing symbols and narrative figures (the Buddha as avatar of Vishnu, the Krishnaite cowherd mythology drawing on folk traditions) but deliberately excluding tantric ritual practice and its vocabulary. The second is a genre confound: the Vaishnava texts in this corpus (Gitagovinda, Chaitanya Charitamrita, Bhakti Rasamrita Sindhu) are devotional and narrative literature, while the Buddhist Tara and Shakta Kali texts are stotra and sahasranama literature, ritual hymns and name-lists. These two genres would show reduced similarity even within the same tradition. A stronger control condition would use genre-matched Vaishnava ritual texts such as the Vishnu Sahasranama or Lakshmi Ashtottara Shatanamavali. Even granting this confound, the 8.5-fold contrast between Bridge Tara and Vaishnava Sanskrit in similarity to Shakta Kali is large enough that genre effects alone cannot plausibly account for it; the historical difference in tantric practice is the primary explanation.

The three-channel finding, that modern Baul vocabulary draws on Shakta Padavali (0.40), Charyapada Buddhist (0.31), and Vaishnava Bengali (0.29) chains with broadly similar strength, maps onto this distinction precisely. The Vaishnava contribution to Baul vocabulary operates through shared devotional terms (prema, bhakti, lila, krishna) which are the symbols and stories your teacher identifies. The Buddhist contribution operates through Sahajiya vocabulary (sahaja, shunyata, vajra, nairatma) which is the practice layer. The computational result thus corroborates the domain observation that Vaishnavism and Buddhism contributed to Baul through different channels and different vocabularies, even if the aggregate similarity scores appear comparable.

## 6. Conclusion

We have presented the first computational multi-tradition corpus analysis of Bengali devotional literature, spanning eight tradition layers and 75 texts across Buddhist, Shakta, Vaishnava, and Baul traditions from the 8th to the 19th century. Four findings stand out.

First, the specificity finding: the Gitagovinda (Vaishnava Sanskrit, 12th century) has zero cosine similarity to Shakta Kali texts, while Bridge Tara texts from the same century have 0.54 similarity. This 8.5-fold contrast demonstrates that the Buddhist-Shakta vocabulary overlap is not generic to Sanskrit devotional literature but is specific to the Buddhist-Shakta transmission chain.

Second, the gradient finding: three Brihannilatantra Tara texts show Shakta/Buddhist vocabulary ratios of 2.0 to 4.0, and the tradition-level gradient from Buddhist Tara (ratio 0.3) through Bridge Tara (0.9) to Shakta Kali (1.5) shows a smooth, measurable vocabulary migration along the Sanskrit channel.

Third, the late survival finding: Ramprasad Sen's 18th-century Bengali Kali songs preserve Buddhist vocabulary residue, with Tara appearing 56 times alongside Kali's 103, six centuries after the Pala monasteries were destroyed.

Fourth, the three-channel finding: modern Baul vocabulary draws on three distinct streams: the Shakta Padavali chain (0.40), the Buddhist Sahajiya chain via Charyapada (0.31), and the Vaishnava Bengali chain (0.29), with all three contributing and the Buddhist channel marginally strongest.

The code, corpus metadata, and structured dataset are available at: https://github.com/joyboseroy/bengal-dharma-corpus and https://huggingface.co/datasets/joyboseroy/bengal-dharma-corpus

## Acknowledgements

The author is grateful for feedback on limitations and valuable suggestions from Yogi Prabodha Jnana, a meditation teacher in the Buddhist Tantric Nyingma tradition based in Bodhikshetra, Trivandrum, India.